\documentclass[letterpaper, 10 pt, conference]{ieeeconf}  

\IEEEoverridecommandlockouts                              

\usepackage{graphics} 
\usepackage{epsfig} 

\usepackage{caption}
\usepackage{subcaption}
\usepackage{float}
\usepackage{multirow}
\usepackage{colortbl,hhline}
\usepackage{amsmath} 
\usepackage{amssymb}  
\usepackage[fleqn,tbtags]{mathtools}
\usepackage{cancel}

\usepackage{color,soul}
\usepackage[bordercolor=white,backgroundcolor=gray!30,linecolor=black,colorinlistoftodos]{todonotes}

\usepackage[linesnumbered,ruled,vlined]{algorithm2e}
\usepackage[noend]{algpseudocode}
\usepackage{booktabs}
\usepackage{tabularx}
\usepackage{tabulary}
\usepackage{graphicx} 
\usepackage[export]{adjustbox}
\usepackage{cite}
\usepackage[super]{nth}
\usepackage{comment}
\let\labelindent\relax
\usepackage{enumitem}
\usepackage{framed}
\usepackage{makecell}
\usepackage{diagbox}

\newcommand{\ie}{{\em i.e.,~}}

\title{\LARGE \bf
FORTE: Forecasting Occupancy for Spatiotemporal Risk-Aware Planning in Dynamic Environments
}

\author{Hahjin Lee and Young J. Kim
\thanks{The authors are with the Department of Computer Science and Engineering at Ewha Womans University in Korea
   ${\it \{hahjinlee|kimy\}@ewha.ac.kr}$.}%
}

\begin{document}

\bstctlcite{BSTcontrol}
\maketitle
\thispagestyle{empty}
\pagestyle{empty}

\begin{abstract}
Safe navigation in dynamic environments requires anticipating future environmental states to account for spatiotemporal risks, specifically {\em when} and {\em where} collisions may occur. To this end, occupancy grid map (OGM) prediction has been widely adopted as an effective approach. However, existing OGM-based navigation methods often struggle to achieve accurate and efficient forecasting and fail to fully exploit the temporal information in predicted OGMs during planning. To address these challenges, we propose FORTE, a navigation framework that directly exploits the spatiotemporal evolution of predicted occupancy from the perspectives of spatiotemporal occupancy overlap and occupancy directivity. Based on these properties, FORTE evaluates multiple topology-distinct paths and selects the suitable one without explicit object detection or tracking. To support online planning, we formulate a latent diffusion model-based OGM predictor that generates the entire forecast horizon in a non-autoregressive manner while maintaining temporal consistency through temporal shift modules. Extensive evaluations demonstrate that FORTE  outperforms state-of-the-art baselines. For prediction, FORTE achieves up to 215.3\% higher IoU and 5.24$\times$ faster inference; for navigation, it yields up to a 3.5$\times$ higher success rate.

\end{abstract}

\section{Introduction} \label{section:intro}
Safe and efficient navigation through spaces shared with moving agents is essential for autonomous mobile robots to perform a wide range of tasks in everyday environments \cite{lee2021service}. However, in such dynamic environments, the surroundings continuously evolve, making collision risks vary across space and time. Accordingly, safe dynamic navigation requires anticipating future environmental states to identify \emph{when} and \emph{where} collisions may occur during planning, enabling proactive collision avoidance.


Consequently, many studies predict the future states of surrounding obstacles and incorporate them into robot path planning. These methods can be categorized into {\em object-level} approaches that forecast future trajectories of individual moving obstacles and {\em scene-level} approaches that model the evolution of the entire scene. Although object-level methods can easily incorporate prediction results into path planning as constraints \cite{wang2022group,samavi2025sicnav}, they rely on complex multi-stage pipelines (e.g., detection and tracking), through which errors can propagate and degrade overall performance. To address these limitations, occupancy grid map (OGM) prediction has been extensively studied as a scene-level approach \cite{schreiber2021dynamic, toyungyernsub2021double, lange2021attention, xie2023stochastic}. OGM forecasting predicts the evolution of occupied regions directly on a grid without explicit object detection or tracking. However, for online planning, recent OGM forecasting methods still struggle to achieve both high prediction quality and efficient inference, while their autoregressive formulation, which feeds predicted results back into the model, further reduces inference efficiency \cite{xie2023stochastic,lei2025diffogmp}.   

Furthermore, dynamic navigation requires not only accurate forecasting but also effective integration of the predicted occupancy information into planning. However, existing OGM prediction-based robot navigation methods commonly use OGMs as a static 2D representation during planning \cite{xie2025scope}, thereby losing the temporal evolution of occupancy. As a result, the planner may find overly conservative or suboptimal path decisions because it cannot account for when a location will be occupied or whether occupancy is approaching or receding. These methods also typically provide a single high-level path, commonly generated by a global planner, for the local planner to follow. This can confine the local planner to a local optimum on the unfavorable side of a moving obstacle, leading to the freezing robot problem \cite{trautman2010unfreezing}.

To address these challenges, we propose FORTE, a navigation framework that incorporates the spatiotemporal evolution of predicted OGMs into trajectory planning. FORTE adopts a hierarchical planning pipeline that consists of a topology-driven spatiotemporal risk-aware (TSR) planner and a local planner. To explore different navigation options, the TSR planner generates multiple geometrically feasible, topology-distinct paths and evaluates their risk using predicted OGMs, considering occupancy at the robot's estimated arrival times and the directivity of predicted occupancy, and selects the suitable path to guide the local planner. In addition, efficient forecasting enables this guidance decision to be updated online. FORTE uses an LDM \cite{rombach2022high} for accurate occupancy forecasting, treating the future OGM sequence as a video and generating all future steps non-autoregressively, while temporal shift modules \cite{an2023latent} maintain motion consistency across prediction steps.

To evaluate the effectiveness of FORTE, we conducted experiments on prediction and navigation tasks. We compared FORTE with state-of-the-art OGM prediction methods \cite{xie2025scope} on both tasks, a video generation method \cite{lu2024vdt} on prediction, and a DRL-based method \cite{liu2024height} on navigation, demonstrating superior performance across both tasks. We further conducted an ablation study to validate the effectiveness of the TSR planner. In summary, the contributions of our work are:

\begin{itemize}
\item We present FORTE, a navigation framework that directly integrates the spatiotemporal evolution of predicted OGMs into guidance trajectory planning over topology-distinct paths, without explicit object detection or tracking.
\item We introduce a TSR planner with arrival-time and directivity costs that evaluate not only whether a path region will be occupied when the robot arrives, but also whether it is becoming occupied or free by distinguishing approaching from receding occupancy.
\item We formulate an LDM-based OGM predictor in a non-autoregressive manner, generating the entire forecast horizon at once to support online planning.
\item We demonstrate that FORTE improves IoU for OGM prediction by up to 215.3\% and accelerates inference by up to 5.24$\times$, while achieving up to a 3.5$\times$ higher success rate compared to state-of-the-art baselines.
\end{itemize}

\section{Related Work}

\subsection{Navigation in Dynamic Environments}
To enable robots to safely operate among moving obstacles, classical approaches, such as ORCA \cite{alonso2013optimal} and MPC-based methods \cite{jian2023dynamic}, have been widely used. With advances in learning-based planning, neural network-based approaches using transformers \cite{wang2024navformer}, diffusion models \cite{mizuta2024cobl}, and deep reinforcement learning \cite{xie2023drl,liu2024height} have been explored for navigation. These approaches provide effective obstacle avoidance but remain limited in proactive planning for future changes.
Accordingly, recent navigation methods have incorporated predictions of future environmental states into planning. Object-level approaches \cite{wang2022group,samavi2025sicnav} incorporate predicted trajectories of moving obstacles as planning constraints. Scene-level approaches \cite{xie2025scope} incorporate future environmental states into planning without explicitly modeling individual obstacles, yet they often fail to exploit temporal information. Recently, world models \cite{zhangresworld} have been used for joint forecasting and planning in autonomous driving. However, these approaches remain limited in their practical integration into robot navigation due to their high computational cost and reliance on vision-centric settings.

\subsection{Occupancy Grid Map Prediction}
OGM prediction has been widely used to effectively anticipate future changes in dynamic scenes. Early studies \cite{schreiber2020motion, schreiber2021dynamic} employed ConvLSTM-based architectures for occupancy prediction. However, ConvLSTM-based methods suffered from blurred predictions and the loss of dynamic objects. To address these limitations, \cite{toyungyernsub2021double} introduced a double-prong ConvLSTM to separately model static and dynamic regions, while \cite{lange2021attention} utilized self-attention mechanisms. In addition, \cite{song20192d, mahjourian2022occupancy, liu2025let} leveraged flow information to improve OGM prediction accuracy. More recently, camera-based approaches have been studied for 3D \cite{leng2025occupancy} and 4D \cite{mohan2026forecastocc} occupancy forecasting in autonomous driving.

Despite these advances, the aforementioned approaches are deterministic and therefore cannot model the uncertainty of future states, an essential capability for reliable navigation in dynamic environments. To address this limitation, recent studies have explored stochastic occupancy prediction. \cite{LangeB-RSS-25} combined a transformer with a VAE-GAN for multi-future occupancy forecasting, while \cite{wang2025diffusion} employed diffusion models for stochastic 4D occupancy prediction. However, these methods are designed for autonomous driving scenarios and are difficult to deploy on resource-constrained mobile robots. For mobile robot navigation, \cite{xie2023stochastic, xie2025scope} employed a VAE-based predictor but suffered from blurry predictions. \cite{lei2025diffogmp} adopted a diffusion model for prediction accuracy, but may incur high inference latency. 
Moreover, these approaches reduce future occupancy to spatial costs for planning, thereby overlooking its temporal evolution. FORTE instead incorporates both the spatial and temporal evolution of predicted occupancy into the planning cost.

\subsection{Topology-Driven Planner}
Topology-driven planners have been widely used to explore distinct navigation behaviors to avoid deadlocks in dynamic environments. \cite{cao2019dynamic} derives a topology-based path by identifying traversable gaps within moving crowds, while \cite{mavrogiannis2023winding} selects trajectories that maximize passing progress based on the winding number. Since the feasibility of paths around moving obstacles depends on temporal information, \cite{de2023globally, de2025topology} generate candidate trajectories in space-time by estimating pedestrian motion under a constant-velocity assumption. Unlike these approaches, our method integrates stochastic occupancy forecasting with topology-driven planning, enabling paths to account for predicted environmental changes and uncertainty.

\begin{figure*}[ht!]
{\includegraphics[width=\textwidth]{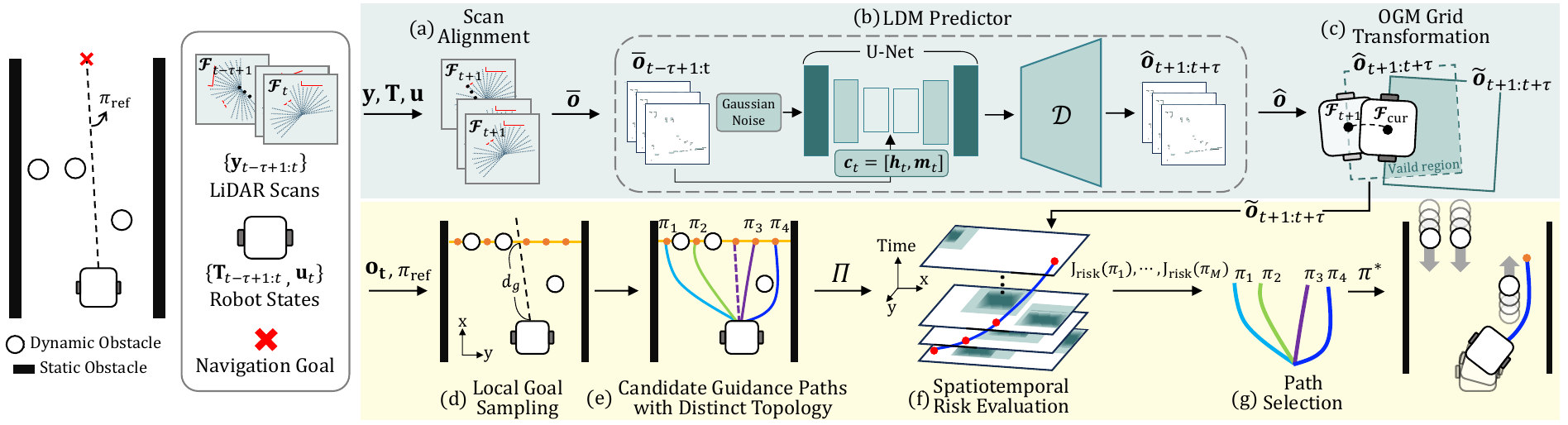}}
\captionof{figure}{ \textbf{FORTE Pipeline.} 
{(a–c) OGM forecasting: At inference, recent LiDAR scans and robot states are used to forecast OGMs using LDM, aligned with the current frame, yielding  $\tilde{\mathbf{o}}_{t+1:t+\tau}$ for planning. (d–g) TSR planning: Given the current OGM $\mathbf{o}_t$ and a reference path $\pi_{\mathrm{ref}}$, the planner generates topology-distinct candidate guidance paths and evaluates their arrival-time and directivity costs using the predicted occupancy evolution. The selected path $\pi^*$ guides the local planner.}}
\label{fig:pipeline}
\vspace{-1.8em}
\end{figure*}

\section{Problem Formulation}
We consider a mobile robot operating in an environment containing both static and dynamic obstacles. At each time step $t$, the robot has a planar pose $\mathbf{T}_t\in SE(2)$, velocity $\mathbf{u}_t=[v_t,\omega_t]^{\mathsf T}$, and a LiDAR scan $\mathbf{y}_t$, from which a binary OGM $\mathbf{o}_t\in\{0,1\}^{C\times H\times W}$ is constructed. Here, $C$ is the number of occupancy channels, and $H$ and $W$ are the height and width of the grid map, respectively. The most recent $\tau$ observations are used to predict the next $\tau$ OGMs at intervals of $\Delta t$, covering a temporal horizon of $\tau\Delta t$.\footnote{In our implementation, $C=1$, $H=W=64$, $\tau=10$, $\Delta t = 0.1\,\mathrm{s}$ and each cell covers $0.1\,\mathrm{m}\times0.1\,\mathrm{m}$.} Each observation is initially expressed in the robot's local frame at the corresponding time step. For planning, the robot is given a navigation goal, and planning is performed online at each time step using the current OGM $\mathbf{o}_t$ and the predicted OGMs $\tilde{\mathbf{o}}_{t+1:t+\tau}$. We formulate the OGM prediction problem in Sec.~\ref{subsection:pf_pred} and Fig.~\ref{fig:pipeline}(a)--(c), and the subsequent planning problem in Sec.~\ref{subsection:pf_plan} and Fig.~\ref{fig:pipeline}(d)--(g).

Our notation distinguishes OGMs at different stages of the observation–prediction–planning pipeline. $\bar{\mathbf{o}}$ denotes past observed OGMs aligned to a common prediction frame, $\hat{\mathbf{o}}$ denotes future OGMs predicted in the same frame, and $\tilde{\mathbf{o}}$ denotes the predictions transformed to the current robot frame for planning.


\subsection{OGM Prediction} \label{subsection:pf_pred}
Since the scans $\mathbf{y}_{t-\tau+1:t}$ are expressed in different local frames $\mathcal{F}_{t-\tau+1:t}$, they are first aligned to a common reference frame for prediction $\hat{\mathcal{F}}_{t+1}$, as described in Sec.~\ref{subsection:pre}. Let $\bar{\mathbf{o}}_{t-\tau+1:t}$ denote the aligned OGM sequence constructed from the aligned scans. The future OGM prediction problem is then formulated as
\begin{equation}
\hat{\mathbf o}_{t+1:t+\tau}
\sim
p_{\theta}\!\left(
\mathbf o_{t+1:t+\tau}
\mid
\bar{\mathbf o}_{t-\tau+1:t}
\right),
\label{eq:ogm_prediction}
\end{equation}
where $p_\theta$ is the learned model in Sec.~\ref{subsection:ldm} and $\hat{\mathbf o}_{t+1:t+\tau}$ is the predicted OGM sequence represented in the frame $\hat{\mathcal{F}}_{t+1}$. Before planning, $\hat{\mathbf o}_{t+1:t+\tau}$ is transformed into the current planning frame $\mathcal{F}_{\mathrm{cur}}$, as described in Sec.~\ref{subsection:post}.

\subsection{TSR Planning} \label{subsection:pf_plan}
We assume that the robot is provided with a reference path $\pi_{\mathrm{ref}}$ toward the navigation goal, for instance, by a global planner. Given $\pi_{\mathrm{ref}}$ and the current OGM $\mathbf{o}_t$, let $\Pi=\{\pi_1,\ldots,\pi_M\}$ denote a set of $M$ geometrically feasible, topology-distinct candidate {\em guidance paths} toward short-horizon local goals sampled laterally at a lookahead distance $d_g$ along $\pi_{\mathrm{ref}}$. 
Given the predicted OGM sequence $\tilde{\mathbf{o}}_{t+1:t+\tau}$, we pair each path point $\mathbf{p}_i^j = (x_i^j,y_i^j) \in \mathbb{R}^2$, which denotes the $j$-th point along a candidate path $\pi_i \in \Pi$, with the robot's estimated arrival time at that point and evaluate the resulting space-time path against $\tilde{\mathbf{o}}_{t+1:t+\tau}$ using the spatiotemporal risk $J_{\mathrm{risk}}$.
The guidance path $\pi^*$ is finally determined by optimizing $J_{\mathrm{risk}}$ while minimizing unnecessary switching between topology classes across planning cycles. The construction of $\Pi$, the definition of $J_{\mathrm{risk}}$, and the path selection procedure are described in Sec.~\ref{section:tsrplanner}. The resulting path $\pi^*$ is provided to a local planner, which handles the robot motion.

\section{OGM Prediction}
This section describes how observations acquired by a moving robot are converted into a future OGM sequence for use in the planning cycle, which consists of three components: aligning past scans, forecasting future occupancy, and transforming predicted OGMs into the current robot frame.

\subsection{Scan Alignment} \label{subsection:pre}

Given past LiDAR scans $\mathbf{y}_{t-\tau+1:t}$, the scans are represented in different local frames $\mathcal{F}_{t-\tau+1:t}$ due to robot motion. If these scans are combined without alignment, even stationary structures may appear to move, so frame inconsistency must be resolved before forecasting. Prior methods align the scans separately to each estimated future frame $\hat{\mathcal F}_{t+1},\ldots,\hat{\mathcal F}_{t+\tau}$, obtained with a constant-velocity motion model \cite{xie2023stochastic}. However, repeating the alignment increases computational cost, while estimation errors over a long horizon can lead to spatial misalignment between $\hat{\mathbf{o}}_{t+1:t+\tau}$ and the environment, potentially degrading planning performance.

FORTE instead estimates only the one-step-ahead frame $\hat{\mathcal F}_{t+1}$ and transforms every scan in $\mathbf y_{t-\tau+1:t}$ into the common frame $\hat{\mathcal F}_{t+1}$ using the robot poses $\mathbf T_{t-\tau+1:t}$, as illustrated in Fig.~\ref{fig:pipeline}(a). All future OGMs are then predicted in $\hat{\mathcal F}_{t+1}$, avoiding repeated scan alignment across future steps and reducing dependence on long-horizon frame estimates.

\subsection{LDM Predictor} \label{subsection:ldm}

To achieve stochastic prediction with efficient inference, we employ an LDM \cite{rombach2022high}. Given the aligned OGMs $\bar{\mathbf{o}}_{t-\tau+1:t}$, we treat the OGM sequence as a short video and generate all $\tau$ future steps non-autoregressively, without recursively using previous predictions as inputs.
Although this formulation improves inference efficiency and avoids error propagation, it may fail to preserve motion consistency across future steps. Inspired by temporal modeling in video generation, we incorporate temporal shift modules \cite{an2023latent} into the U-Net to exchange a subset of latent feature channels between adjacent future steps.

The LDM predictor is conditioned on static and dynamic information. From $\bar{\mathbf{o}}_{t-\tau+1:t}$, a Bayesian occupancy update produces a static map $\mathbf m_t$ \cite{thrun2003learning}, representing persistently observed structures over the past $\tau$ steps, while a ConvLSTM extracts motion feature $\mathbf h_t$ representing recent occupancy changes \cite{xie2025scope}. Their concatenation $\mathbf{c}_t=[\mathbf{h}_t,\mathbf{m}_t]$ conditions the LDM. During training, the target future OGM sequence is encoded into the latent space, where the U-Net is optimized using $\mathbf{v}$-parameterization \cite{salimans2022progressive}. 
At inference, as shown in Fig.~\ref{fig:pipeline}(b), Gaussian noise is denoised by the U-Net conditioned on $\mathbf{c}_t$, and the resulting latent sequence is passed through the decoder $\mathcal D$ to obtain $\hat{\mathbf o}_{t+1:t+\tau}$.  
Overall, the proposed predictor enables accurate, efficient, and temporally consistent OGM forecasting for online planning under the standard diffusion objective.

Since the TSR planner requires a single OGM sequence on which to score paths, FORTE averages the stochastic predictions as $\left[\hat{\mathbf{o}}_{t+k}\right]_{(h,w)}=\frac{1}{N_s}\sum_{n=1}^{N_s}\left[\hat{\mathbf{o}}_{t+k}^{(n)}\right]_{(h,w)}$ for $k=1,\ldots,\tau$. $N_s$ is the number of stochastic samples and $(h,w)$ indexes a grid cell, and $k$ denotes the future prediction step.

\subsection{OGM Grid Transformation} \label{subsection:post}
The fixed prediction frame enables efficient forecasting but becomes misaligned with the current robot frame as the robot may move during inference. Before each planning cycle, we therefore transform $\hat{\mathbf o}_{t+1:t+\tau}$ from $\hat{\mathcal F}_{t+1}$ into the current robot frame $\mathcal F_{\mathrm{cur}=t+\epsilon}$, as illustrated in Fig.~\ref{fig:pipeline}(c). Occupancy probabilities are converted to logits, bilinearly interpolated under the relative frame transformation, converted back to probabilities, with cells outside the valid overlap set to zero. The resulting sequence $\tilde{\mathbf o}_{t+1:t+\tau} $ preserves the forecast in $\hat{\mathbf o}_{t+1:t+\tau}$ while spatially aligning it with the environment used by the TSR planner. The next section explains how the TSR planner incorporates this predicted sequence into the planning problem.

\section{Topology-Driven \\Spatiotemporal Risk Aware Planner} \label{section:tsrplanner}
Accurate OGM forecasting provides useful information for planning, but planning that does not explicitly account for temporal occupancy changes may lead to inappropriate decisions \cite{xie2025scope}. To exploit these changes, we propose the TSR planner, which considers multiple candidate guidance paths with distinct topologies around occupied regions. These paths represent \emph{where} the robot can progress, while the predicted OGM sequence indicates \emph{when} occupancy is expected. Based on this spatiotemporal information, the TSR planner selects the path topology that best aligns with the predicted occupancy evolution, helping the local planner avoid unfavorable local optima. The TSR proceeds as follows

\begin{enumerate}

    \item \textbf{Local Goal Sampling:} TSR samples laterally distributed local goals at a lookahead distance $d_g$ along $\pi_{\mathrm{ref}}$, providing diverse local targets while maintaining progress in the goal direction.

    \item \textbf{Candidate Guidance Paths with Distinct Topology:} For each local goal, TSR generates a geometrically feasible path on $\mathbf{o}_t$. One representative path per topology class is retained to form the candidate guidance set.

    \item \textbf{Spatiotemporal Risk Evaluation:} TSR maps candidate guidance paths into a space-time representation using estimated arrival times and computes their risk cost $J_{\mathrm{risk}}$ by combining arrival-time and directivity costs derived from the predicted OGMs $\tilde{\mathbf{o}}_{t+1:t+\tau}$.

    \item \textbf{Path Selection:} TSR selects the guidance path $\pi^*$ based on $J_{\mathrm{risk}}$ and a selection criterion that reduces oscillatory switching between topology classes, and  $\pi^*$ is then followed by the downstream local planner.
\end{enumerate}


\subsection{Local Goal Sampling}
Since $\tilde{\mathbf{o}}_{t+1:t+\tau}$ covers a limited temporal horizon, path risks can only be evaluated within this horizon. Accordingly, path endpoints are restricted to local goals. To provide guidance within the limited horizon while promoting topological diversity, multiple local goals are sampled using a local goal line \cite{zhang2025ga}{, illustrated as the yellow line in Fig.~\ref{fig:pipeline}(d)}. {The line is placed at a lookahead distance $d_g=v_{\max}\tau\Delta t$ along $\pi_{\mathrm{ref}}$, where $v_{\max}$ is the robot's maximum translational speed, covering the maximum travel distance over the prediction horizon. The line extends laterally across $\pi_{\mathrm{ref}}$.} As this line may intersect obstacles, its free-space portions are divided into multiple line segments. Local goals, shown as the orange dots in Fig.~\ref{fig:pipeline}(d), are then uniformly sampled from each segment, with the number of samples proportional to its length. A geometric path is then generated toward each local goal, yielding {\em candidate guidance paths}.

\subsection{Candidate Guidance Paths with Distinct Topology}
Paths generated toward different local goals may differ geometrically but still represent the same navigation decision, such as passing an occupied region on the same side. Evaluating the risk of all such paths would unnecessarily increase the computation. The TSR planner, therefore, performs a fast, approximate topology check, groups the paths by topology class, and retains one representative from each class for subsequent spatiotemporal risk evaluation.

For each local goal, an A* search is performed on $\mathbf{o}_t$ in parallel to generate a geometric path. Let $\mathcal{O}_s \subset \mathcal{O}$ denote a uniformly sampled subset of occupied cells in $\mathbf{o}_t$ used for the topology check. For each pair of paths, two rays are cast from each $c \in \mathcal{O}_s$ in opposite directions parallel to the local goal line. If the two rays first encounter different paths, $c$ is considered to separate the pair, indicating that the paths are routed on opposite sides of the occupied region. Otherwise, the paired paths are considered to belong to the same approximate topology class if no sampled occupied cell separates them. Through this process, paths are efficiently grouped into approximate topology classes, {as illustrated in Fig.~\ref{fig:pipeline}(e), where paths belonging to distinct classes are represented by different colors.} One representative path (solid line) is then retained from each class, yielding $\Pi=\{\pi_1,\ldots,\pi_M\}$ for the following spatiotemporal risk evaluation.

\subsection{Spatiotemporal Risk Evaluation} \label{subsection:risk_eval}

\begin{figure*}[ht!]
{\includegraphics[width=\textwidth]{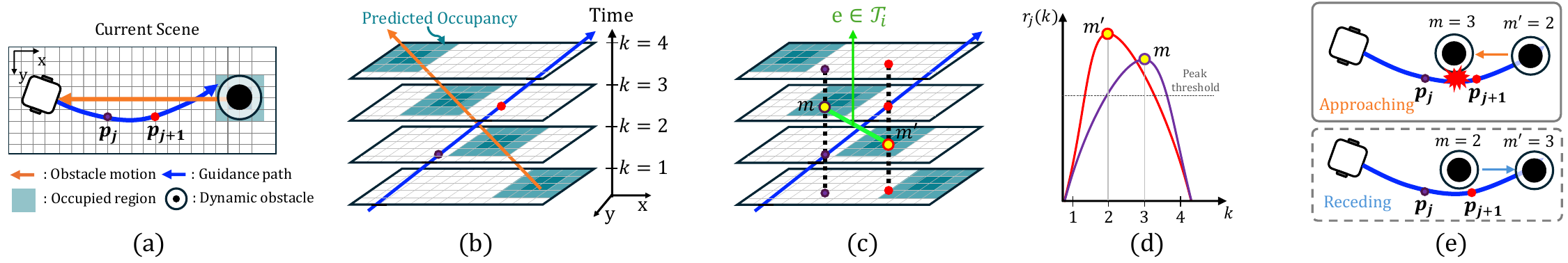}}
\captionof{figure}{
\textbf{Directivity cost from predicted occupancy propagation.} (a) A candidate guidance path and an approaching obstacle. (b) The path points are mapped into the forecasted OGM sequence using estimated arrival times. (c) Occupancy peaks at adjacent path points $\mathbf{p}_j$ and $\mathbf{p}_{j+1}$ are connected to form an edge $e\in\mathcal{T}_i$. (d) Occupancy peaks earlier at the farther point $\mathbf{p}_{j+1}$ ($m'=2$) than at the nearer point $\mathbf{p}_j$ ($m=3$). (e) $m'<m$ indicates that occupancy propagates toward the robot along the path, thereby incurring a positive directivity cost; the reverse ordering is shown in the dashed box.
}
\label{fig:planningpipe2}
\vspace{-1.8em}
\end{figure*}

A geometric path specifies \emph{where} the robot may travel, but not \emph{when} it will reach each point along the path. Since the predicted OGM sequence varies over time, evaluating a path against the forecast requires establishing a temporal correspondence between the path and the prediction. We establish this correspondence by estimating the robot’s arrival time at each path point and assigning the corresponding predicted OGM layer to that point. For each representative path $\pi_i=(\mathbf{p}_i^0,\ldots,\mathbf{p}_i^{N_i})\in \Pi$, let $d_i^j$ denote the cumulative distance from $\mathbf{p}_i^0$ to $\mathbf{p}_i^j$. We estimate the robot's arrival time along the path using $\bar{v}=\max(v_t,v_{\min})$, where $v_t$ is the robot’s current translational speed. The lower speed bound $v_{\min}$ keeps the estimate defined when the robot is nearly stationary. The estimated arrival time is then computed as $a_i^j=d_i^j/\bar{v}$, and the corresponding index of the predicted OGM sequence is defined as $k_i^j=\min\{\tau,\max\{1,\lfloor a_i^j/\Delta t +1/2\rfloor\}\}$. The index is clipped to remain within the prediction horizon. {As illustrated in Fig.~\ref{fig:pipeline}(f), each $\mathbf{p}_i^j$ is thus paired with an OGM layer, represented by a red dot, providing a space-time representation of the path.} Each $\pi \in \Pi$ is then evaluated based on an {\em arrival-time cost} and a {\em directivity cost} using $\tilde{\mathbf{o}}_{t+1:t+\tau}$. For brevity, $\tilde{\mathbf{o}}_{t+k}$ is denoted by $\tilde{\mathbf{o}}_k$ hereafter.

\subsubsection{Arrival-Time Cost}
Spatial overlap between a path and predicted occupancy does not necessarily imply a collision risk, since the robot and the occupancy may reach the same location at different times. We therefore define the arrival-time cost of each path as
\begin{equation}
J_{\mathrm{arr}}(\pi_i)
=
\sum_{j=1}^{N_i}
\tilde{\mathbf{o}}_{k_i^j}(\mathbf{p}_i^j).
\label{eq:arrivalcost}
\end{equation}
$\tilde{\mathbf{o}}_{k_i^j}(\mathbf{p}_i^j)$ denotes the occupancy cost at $\mathbf{p}_i^j$ in the $k_i^j$-th OGM layer. This cost penalizes a path only when its points are predicted to be occupied when the robot reaches them, avoiding overly conservative rejection of paths that remain feasible in space and time.

\subsubsection{Directivity Cost}
The arrival-time cost evaluates only the occupancy cost at each path point when the robot reaches it. Thus, even when actual collision risk increases as an occupied region approaches the robot, it may be represented by only a single occupancy cost at the time when the robot and the occupied region overlap along the path, which may fail to reflect the increasing risk. To complement the arrival-time cost, the directivity cost considers the direction of predicted occupancy motion. The directivity cost captures this temporal trend without explicit object tracking by examining the temporal order of {\em occupancy peaks} along the path. An occupancy peak represents a time at which a moving obstacle is predicted to strongly occupy the spatial location of a path point. Along the path, if occupancy peaks occur earlier (\ie at smaller OGM-layer indices) at locations farther from the robot and progressively later (\ie at larger OGM-layer indices) at locations nearer to the robot, the occupied region is interpreted as approaching the robot. The reverse ordering indicates receding motion.

For simplicity, we abbreviate $\mathbf{p}_j = \mathbf{p}_i^j$ for a fixed path $\pi_i$. Along $\pi_i$, let ${r}_{j}(k)=\tilde{\mathbf{o}}_{k}(\mathbf{p}_j)$ denote the occupancy cost at $\mathbf{p}_j$ at prediction step $k$ (\ie the $k$-th OGM layer). The sequence $\{r_j(k)\}_{k=1}^{\tau}$ describes how occupancy at $\mathbf{p}_j$ is predicted to evolve over time. Local maxima of $r_j(k)$ across the prediction horizon that exceed a fixed threshold are defined as \emph{occupancy peaks}, and their indices form $\mathcal{P}_j$. To trace the motion of occupied regions within a plausible temporal range, an edge is established between $m \in \mathcal{P}_{j}$ and $m' \in \mathcal{P}_{j+1}$ if $1 \le |m' - m| \le B$, where $B$ is the temporal window size. Each edge $e =((j,m),(j+1,m'))$ belongs to $\mathcal{T}_{i}$, as illustrated in Fig.~\ref{fig:planningpipe2}(c) and (d). Each edge represents the propagation of an occupancy peak from one path point to the next. To compute the directivity cost, each $e$ is characterized by its signed speed and occupancy cost. The signed speed is defined as $v_e = {\|\mathbf{p}_{j+1}-\mathbf{p}_{j}\|_2}/ {((m'-m)\Delta t)}$. In addition, occupancy cost is defined as $\rho_e=(r_j(m)+r_{j+1}(m'))/2$. The directivity score is defined as 
\begin{equation}
    J_{\text{dir}}(\pi_{i}) = \sum_{e \in \mathcal{T}_{i}} \rho_e[-\tanh(v_e)].
\label{eq:directivitycost}
\end{equation}
Since $\mathbf{p}_j$ is closer to the robot than $\mathbf{p}_{j+1}$, $m'<m$ yields $v_e<0$, increasing the cost for approaching occupancy, whereas $m'>m$ yields $v_e>0$, reducing the cost for receding occupancy. It enables proactive and safe planning by favoring paths through spaces to be vacated while avoiding regions expected to be occupied by approaching obstacles. {The justification of  $J_{\mathrm{dir}}$ is empirically validated by the ablation results in Sec.~\ref{subsection:sim_exp} (Table~\ref{tab:sim_navigation}).}

\subsubsection{Total Cost}
The total spatiotemporal risk cost for $\pi_i$, combining arrival-time and directivity costs, is defined as
\begin{equation}
J_{\mathrm{risk}}(\pi_i)
=
w_{\mathrm{arr}}J_{\mathrm{arr}}(\pi_i)
+
w_{\mathrm{dir}}J_{\mathrm{dir}}(\pi_i).
\end{equation}
where $w_{\mathrm{arr}}$ and $w_{\mathrm{dir}}$ control the relative contributions of the arrival-time and directivity costs, respectively. 

\begin{figure*}[ht!]
{\includegraphics[width=\textwidth]{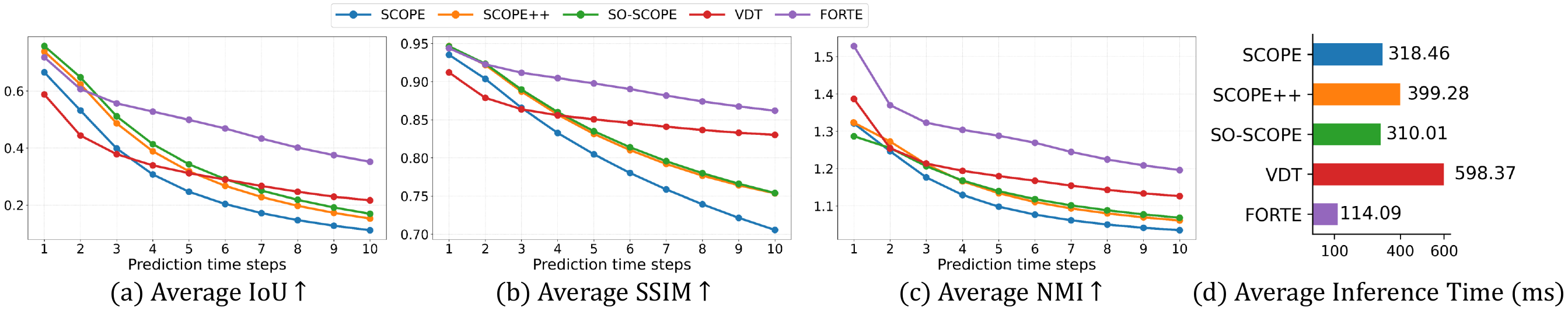}}
\captionof{figure}{\textbf{Prediction performance.} Average IoU, SSIM, and NMI at each of the 10 prediction steps, averaged over the test set. The curves show the mean over 8 stochastic prediction samples.}
\label{fig:pred_result}
\vspace{-1.8em}
\end{figure*}
\vspace{-0.5em}
\subsection{Path Selection}
After risk evaluation, the guidance path is selected by minimizing $J_{\mathrm{risk}}$. However, selecting the path with the minimum $J_{\mathrm{risk}}$ at each planning cycle can lead to oscillatory switching between topology classes. TSR therefore employs a path selection criterion that retains the previous guidance path and its topology as long as the associated risk remains acceptable, but allows a different route to be selected otherwise.

Let $\pi_{\mathrm{prev}}$ denote the guidance path selected in the previous planning cycle. It is pruned and extended to the nearest current local goal, producing $\pi_{\mathrm{prev}}^*$. If $J_{\mathrm{risk}}(\pi_{\mathrm{prev}}^*)\leq\eta$, where $\eta$ is a predefined risk threshold, the TSR planner reuses $\pi_{\mathrm {prev}}^*$, thereby reducing replanning overhead. Otherwise, the planner generates topology-distinct paths and first considers the path $\pi_h$ that belongs to the same topology class as $\pi_{\mathrm{prev}}$. If $J_{\mathrm{risk}}(\pi_h)\leq\eta$, $\pi_h$ is selected; otherwise, $\pi$ with $\arg\min_{\pi_i\in\Pi}J_{\mathrm{risk}}(\pi_i)$ is selected. Finally, the selected path $\pi^*$ is provided to the local planner as guidance path.

\section{Experiments}\label{sec:exp}

To demonstrate the prediction and navigation performance of FORTE, we first evaluate our method on a dataset and then validate its navigation performance in both simulated and real-world environments.
\subsection{Experimental Setup}
\textbf{Implementation:} The prediction and simulated navigation experiments are conducted on a system equipped with an AMD Ryzen 5 3600 CPU and an NVIDIA GeForce RTX 3090 GPU, running Ubuntu 20.04 and ROS Noetic.

\textbf{Baselines:} We compare FORTE with several baselines. 

\begin{enumerate}
    \item \textbf{SCOPE-based methods}~\cite{xie2025scope}: SCOPE, SCOPE++, and SO-SCOPE are stochastic OGM prediction-based navigation methods that use occupancy predictions for planning. \textbf{SCOPE} introduces a stochastic prediction framework,  \textbf{SCOPE++} extends it by incorporating static map conditioning, and \textbf{SO-SCOPE} further improves computational efficiency through knowledge distillation and uses statistical uncertainty estimation.
    
    \item \textbf{VDT}~\cite{lu2024vdt}: Diffusion-based video prediction method with spatial and temporal attention. We replace the LDM predictor in our framework with VDT and train it with the OGM-Turtlebot2 dataset \cite{xie2023stochastic}.
    
    \item \textbf{HEIGHT}~\cite{liu2024height}: DRL-based navigation method that models heterogeneous interactions among humans, robots, and obstacles. HEIGHT is fine-tuned under our experimental setting, with human positions detected as described in HEIGHT. 
\end{enumerate}

For the OGM prediction experiments, FORTE is compared with SCOPE-based methods and VDT. VDT is excluded from the navigation experiments due to its high inference latency, which limits its applicability to real-time navigation. For the navigation experiments in simulation, we compare SCOPE-based methods and HEIGHT with FORTE integrated with two downstream local planners, DWA\cite{fox2002dynamic} and ART-TEB\cite{lee2025adaptive}.

\subsection{OGM Prediction Results} \label{subsection:pred_exp}
We evaluate all prediction methods on 17,000 test samples over 10 future prediction steps. The evaluation metrics include IoU measuring occupancy overlap, SSIM evaluating structural similarity, and NMI evaluating the similarity of occupancy distributions. We use the OGM-Turtlebot2 dataset \cite{xie2023stochastic}, collected with a Turtlebot2 in a dynamic environment and consisting of robot states and LiDAR scans.

\textbf{Performance Comparison} 
As shown in Fig.~\ref{fig:pred_result}(a) and (b), FORTE achieves competitive performance for the short prediction horizon and consistently outperforms the other methods for the longer horizon, with improvements of up to 215.3\% in IoU and 22.14\% in SSIM. This advantage at longer horizons can be attributed to the proposed architecture, which captures spatiotemporal dependencies and compensates for robot ego-motion to maintain spatial alignment.
As shown in Fig.~\ref{fig:pred_result}(c), the diffusion-based methods, VDT and FORTE, generally achieve higher NMI values, with FORTE outperforming all competing methods. Since NMI reflects the occupancy probability distribution, the higher values suggest that diffusion-based prediction produces cleaner distributions through iterative denoising.

\textbf{Inference Time Comparison}
We further compare the computational efficiency of the prediction methods. As shown in Fig.~\ref{fig:pred_result}(d), FORTE achieves the lowest inference time among all methods, being 2.79$\times$, 3.50$\times$, and 2.72$\times$ faster than SCOPE, SCOPE++, and SO-SCOPE, respectively. In particular, compared with the VDT, FORTE achieves approximately 5.24$\times$ faster inference. This efficiency is achieved by performing diffusion in the latent space and predicting the entire OGM sequence non-autoregressively, enabling efficient prediction for online navigation.


\begin{figure*}[ht!]
    \centering
    \begin{subfigure}[t]{\textwidth}
        \includegraphics[width=\linewidth]{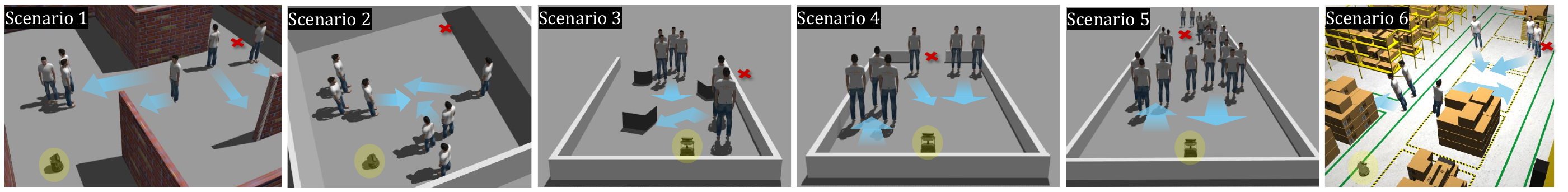}
    \end{subfigure}
    \caption{\textbf{Simulation Scenarios.} Six navigation scenarios with diverse pedestrian motions and spatial constraints. Blue arrows indicate the pedestrian motion directions. Start positions (yellow circles) and goal positions (red crosses) are marked.}
    \label{fig:sim_exp}
    \vspace{-1.0em}
\end{figure*}

\begin{table*}[t]
\renewcommand{\arraystretch}{0.9}
\setlength{\tabcolsep}{3.2pt}
\centering
\scriptsize

\resizebox{\textwidth}{!}{%
\begin{tabular}{c|c|cc|cc|cc|cc|cc|cc}
\toprule[1.5pt]

\multirow{2}{*}[-2pt]{\makecell[c]{Local\\Planner}}
& \multirow{2}{*}[-2pt]{Method}
& \multicolumn{2}{c|}{Scenario 1}
& \multicolumn{2}{c|}{Scenario 2}
& \multicolumn{2}{c|}{Scenario 3}
& \multicolumn{2}{c|}{Scenario 4}
& \multicolumn{2}{c|}{Scenario 5}
& \multicolumn{2}{c}{Scenario 6} \\

\cmidrule(lr){3-4}
\cmidrule(lr){5-6}
\cmidrule(lr){7-8}
\cmidrule(lr){9-10}
\cmidrule(lr){11-12}
\cmidrule(lr){13-14}

& & SR (\%) & TTG (s) & SR (\%) & TTG (s) & SR (\%) & TTG (s) & SR (\%) & TTG (s) & SR (\%) & TTG (s) & SR (\%) & TTG (s) \\

\midrule

\multirow{5}{*}{DWA}
& SCOPE & 14 & 26.1 & 56 & 22.1 & 16 & 30.55 & 18 & 21.63 & 22 & 49.32 & 32 & 31.4 \\
& SCOPE++ & 30 & 28.0 & 72 & 21.71 & 6 & 30.13 & 60 & 21.82 & 8 & 56.52 & 2 & 33 \\
& SO-SCOPE & 22 & 30.3 & 60 & 22.8 & 6 & 30.03 & 48 & 22.92 & 0 & -- & 12 & 31.2 \\
& \makecell[c]{FORTE w/o TSR planner} & 44 & 22.86 & 74 & 20.36 & 58 & 25.87 & 66 & 20.24 & 22 & 52.78 & 72 & 29.74 \\
& \makecell[c]{FORTE w/o Directivity cost} & 82 & 19.4 & 84 & 16.72 & 68 & 19.21 & 68 & 18.12 & 30 & 42.1 & 76 & 23.87 \\
& FORTE & 84 & 21.0 & \bf{94} & \bf{15.63} & \bf{82} & \bf{17.6} & 90 & \bf{14.38} & \bf{80} & \bf{39.56} & \bf{88} & 24.3 \\
\cmidrule(r){1-2}
ART-TEB & FORTE & \bf{88} & \bf{18.77} & 86 & 15.92 & 80 & 18.95 & \bf{98} & 14.56 & 74 & 42.56 & \bf{88} & \bf{22.56} \\
\cmidrule(r){1-2}
-- & HEIGHT & 20 & 36.1 & 84 & 22.1 & 4 & 33.5 & 4 & 20.45 & 6 & 78.5 & 80 & 30.28 \\
\bottomrule[1.5pt]
\end{tabular}%
}

\caption{\textbf{Navigation performance in Simulation.}
Success Rate (SR) and Time to Goal (TTG) are reported for each scenario; TTG is for successful trials only.}
\label{tab:sim_navigation}
\vspace{-2.8em}
\end{table*}

\subsection{Navigation Results in Simulation} \label{subsection:sim_exp}
As shown in Fig.~\ref{fig:sim_exp}, the simulation experiments were conducted in Gazebo using a TurtleBot2 across six scenarios, with 50 trials per scenario. All methods used the same start and goal positions and identical parameters for the same planner. At each planning cycle, all OGM prediction-based methods used 8 stochastic samples for the planning. For the TSR planner, a straight-line path served as the reference path $\pi_{\mathrm{ref}}$. Following \cite{xie2025scope}, the SCOPE-based methods used a conventional hierarchical planning pipeline in which an A* planner generated a path to the navigation goal and a local planner followed it. In FORTE w/o TSR planner, the A* planner was used in place of the TSR planner.
 
\textbf{Comparison} 
As summarized in Table~\ref{tab:sim_navigation}, FORTE-based methods outperformed the baselines in both success rate (SR) and time to goal (TTG), demonstrating robust navigation performance. 
In terms of SR, FORTE with DWA and ART-TEB achieved average SRs of 86.3\% and 85.7\%, respectively, across the six scenarios. In comparison, SCOPE, SCOPE++, SO-SCOPE, and HEIGHT achieved average SRs of 26.3\%, 29.7\%, 24.7\%, and 33.0\%, respectively. Thus, the FORTE-based methods achieved $2.6\times \sim 3.5\times$ higher average SRs than the baselines. 
In Scenario 6, SR substantially differs across methods because successful navigation is particularly sensitive to timing. Waiting until the space clears is often sufficient, allowing HEIGHT to achieve high SRs through its waiting strategy, whereas SCOPE-based methods often enter before the space is clear, resulting in repeated collisions with oncoming obstacles.
For TTG, FORTE-based methods achieved faster TTGs than the baselines, being up to approximately 2$\times$ faster. The faster TTG is mainly attributed to the planner’s ability to exploit spatiotemporal occupancy, thereby using obstacle motion and selecting routes through regions expected to become free. As a result, unnecessary stops and waiting time are reduced, leading to faster navigation.

We further compare the computation times of the TSR and A* planner used in the baseline navigation, both executed every planning cycle at 10 Hz. Despite involving multiple stages, the TSR planner achieves an average computation time of 22.19 ms, comparable to 22.48 ms for the A* planner. This efficiency is achieved through parallel candidate guidance path generation and short-horizon planning using local goals.

FORTE performs strongly with both DWA and ART-TEB, with DWA outperforming ART-TEB in more scenarios. This can be attributed to ART-TEB requiring a continuous path to be generated under kinodynamic constraints, which becomes more difficult when safe space is limited by moving obstacles, whereas DWA directly selects feasible local velocities. 

\textbf{Ablation Study}
The effect of the TSR planner can be observed from the results of FORTE w/o TSR planner in Table~\ref{tab:sim_navigation}. Without the TSR planner, FORTE uses an A* planner that provides a single path to the local planner. Even in this setting, FORTE achieves higher SR than the other methods in most scenarios, indicating the benefit of accurate OGM prediction for navigation. However, as shown by the results, prediction alone is insufficient to enable effective planning in dynamic environments. This highlights the contribution of the TSR planner.
The benefit of the directivity cost can be also observed from the results of FORTE w/o Directivity cost in Table~\ref{tab:sim_navigation}. Its effect is particularly evident in Scenario 5, which involves bidirectional pedestrian motion along the robot's direction of travel, with pedestrians moving both toward and away from the robot. Without this cost, SR decreases from 80\% to 30\%, while TTG increases.

\begin{figure}[!t]
    \centering
    \includegraphics[width=\columnwidth]{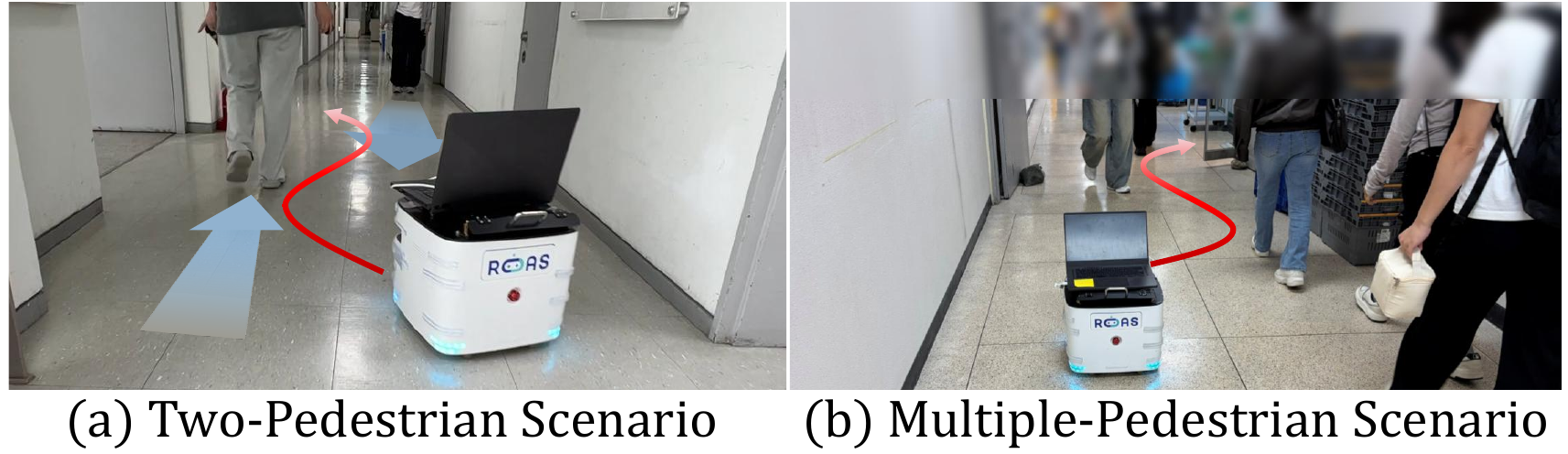}
    \caption{{\textbf{Real-world Scenarios.} (a) Two pedestrians moving in opposite directions. (b) Multiple pedestrians moving in various directions. Both scenarios are set in corridor environments. Red curves indicate the robot paths.}}
    \label{fig:real_exp}
    \vspace{-0.7em}
\end{figure}

\vspace{-1.3em}

\subsection{Navigation Results In Real-world } \label{subsubsection:real_exp}

The real-world experiments were conducted using a differential-drive mobile robot. All computations were processed on an onboard computer (AMD Ryzen 9 5900HS, RTX 3080) mounted on the robot.

FORTE was compared with SO-SCOPE, the most computationally efficient SCOPE-based method and thus the most suitable baseline for real-time navigation, in the two-pedestrian scenario shown in Fig.~\ref{fig:real_exp}(a). Over trials, FORTE (w/ DWA) succeeded in all trials (5/5), whereas SO-SCOPE succeeded in only 2/5. 
{FORTE was additionally demonstrated in a crowded corridor with more than 10 pedestrians moving in different directions, as shown in Fig.~\ref{fig:real_exp}(b), to show its effectiveness in challenging scenarios.}

\section{Conclusion}
In this paper, we presented FORTE to address the challenges of safe navigation in dynamic environments. Extensive prediction and navigation experiments demonstrated improved prediction accuracy and navigation performance, with ablation results supporting the contribution of the TSR planner and directivity cost to navigation success. However, the current framework relies on a conventional local planner, which limits the range of robot motions, such as backward motion. In addition, prediction and planning are coupled through the forecasted OGMs but are not jointly optimized. Future work will investigate the formal analysis of the directivity cost and joint prediction framework that simultaneously predicts future robot motions and OGMs, enabling more flexible planning that accounts for their interactions.





\bibliographystyle{IEEEtran}
\bibliography{main}

@IEEEtranBSTCTL{BSTcontrol,
  CTLdash_repeated_names = "no"
}

@article{lee2021service,
  title={Service robots: a systematic literature review},
  author={Lee, In},
  journal={Electronics},
  volume={10},
  number={21},
  pages={2658},
  year={2021},
  publisher={MDPI}
}

@inproceedings{lei2025diffogmp,
  title={DiffOGMP: Diffusion Model for Stochastic Occupancy Grid Map Prediction in Dynamic Scenes},
  author={Lei, Yunfei and Zhao, Jiarui and Sun, Shengdi and Du, Chenyang and Wu, Xirui and Wang, Jian},
  booktitle={2025 IEEE 23rd International Conference on Industrial Informatics (INDIN)},
  pages={1--8},
  year={2025},
  organization={IEEE}
}

@inproceedings{rombach2022high,
  title={High-resolution image synthesis with latent diffusion models},
  author={Rombach, Robin and Blattmann, Andreas and Lorenz, Dominik and Esser, Patrick and Ommer, Bj{\"o}rn},
  booktitle={Proceedings of the IEEE/CVF conference on computer vision and pattern recognition},
  pages={10684--10695},
  year={2022}
}

@inproceedings{alonso2013optimal,
  title={Optimal reciprocal collision avoidance for multiple non-holonomic robots},
  author={Alonso-Mora, Javier and Breitenmoser, Andreas and Rufli, Martin and Beardsley, Paul and Siegwart, Roland},
  booktitle={Distributed autonomous robotic systems: The 10th international symposium},
  pages={203--216},
  year={2013},
  organization={Springer}
}

@inproceedings{jian2023dynamic,
  title={Dynamic control barrier function-based model predictive control to safety-critical obstacle-avoidance of mobile robot},
  author={Jian, Zhuozhu and Yan, Zihong and Lei, Xuanang and Lu, Zihong and Lan, Bin and Wang, Xueqian and Liang, Bin},
  booktitle={2023 IEEE international conference on robotics and automation (ICRA)},
  pages={3679--3685},
  year={2023},
  organization={Ieee}
}

@article{wang2024navformer,
  title={NavFormer: A transformer architecture for robot target-driven navigation in unknown and dynamic environments},
  author={Wang, Haitong and Tan, Aaron Hao and Nejat, Goldie},
  journal={IEEE Robotics and Automation Letters},
  volume={9},
  number={8},
  pages={6808--6815},
  year={2024},
  publisher={IEEE}
}

@inproceedings{mizuta2024cobl,
  title={Cobl-diffusion: Diffusion-based conditional robot planning in dynamic environments using control barrier and lyapunov functions},
  author={Mizuta, Kazuki and Leung, Karen},
  booktitle={2024 IEEE/RSJ International Conference on Intelligent Robots and Systems (IROS)},
  pages={13801--13808},
  year={2024},
  organization={IEEE}
}

@article{xie2023drl,
  title={Drl-vo: Learning to navigate through crowded dynamic scenes using velocity obstacles},
  author={Xie, Zhanteng and Dames, Philip},
  journal={IEEE Transactions on Robotics},
  volume={39},
  number={4},
  pages={2700--2719},
  year={2023},
  publisher={IEEE}
}

@article{liu2024height,
  title={HEIGHT: Heterogeneous Interaction Graph Transformer for Robot Navigation in Crowded and Constrained Environments},
  author={Liu, Shuijing and Xia, Haochen and Pouria, Fatemeh Cheraghi and Hong, Kaiwen and Chakraborty, Neeloy and Driggs-Campbell, Katherine},
  journal={IEEE Transactions on Automation Science and Engineering},
  year={2026}
}

@inproceedings{wang2022group,
  title={Group-based motion prediction for navigation in crowded environments},
  author={Wang, Allan and Mavrogiannis, Christoforos and Steinfeld, Aaron},
  booktitle={Conference on Robot Learning},
  pages={871--882},
  year={2022},
  organization={PMLR}
}

@article{samavi2025sicnav,
  title={Sicnav-diffusion: Safe and interactive crowd navigation with diffusion trajectory predictions},
  author={Samavi, Sepehr and Lem, Anthony and Sato, Fumiaki and Chen, Sirui and Gu, Qiao and Yano, Keijiro and Schoellig, Angela P and Shkurti, Florian},
  journal={IEEE Robotics and Automation Letters},
  year={2025},
  publisher={IEEE}
}

@inproceedings{zhangresworld,
  title={ResWorld: Temporal Residual World Model for End-to-End Autonomous Driving},
  author={Zhang, Jinqing and Fu, Zehua and Liu, Qingjie and Wang, Yunhong and others},
  booktitle={The Fourteenth International Conference on Learning Representations}
}

@inproceedings{schreiber2020motion,
  title={Motion estimation in occupancy grid maps in stationary settings using recurrent neural networks},
  author={Schreiber, Marcel and Belagiannis, Vasileios and Gl{\"a}ser, Claudius and Dietmayer, Klaus},
  booktitle={2020 IEEE International Conference on Robotics and Automation (ICRA)},
  pages={8587--8593},
  year={2020},
  organization={IEEE}
}

@inproceedings{schreiber2021dynamic,
  title={Dynamic occupancy grid mapping with recurrent neural networks},
  author={Schreiber, Marcel and Belagiannis, Vasileios and Gl{\"a}ser, Claudius and Dietmayer, Klaus},
  booktitle={2021 IEEE International Conference on Robotics and Automation (ICRA)},
  pages={6717--6724},
  year={2021},
  organization={IEEE}
}

@inproceedings{toyungyernsub2021double,
  title={Double-prong convlstm for spatiotemporal occupancy prediction in dynamic environments},
  author={Toyungyernsub, Maneekwan and Itkina, Masha and Senanayake, Ransalu and Kochenderfer, Mykel J},
  booktitle={2021 IEEE International Conference on Robotics and Automation (ICRA)},
  pages={13931--13937},
  year={2021},
  organization={IEEE}
}

@inproceedings{lange2021attention,
  title={Attention augmented convlstm for environment prediction},
  author={Lange, Bernard and Itkina, Masha and Kochenderfer, Mykel J},
  booktitle={2021 IEEE/RSJ International Conference on Intelligent Robots and Systems (IROS)},
  pages={1346--1353},
  year={2021},
  organization={IEEE}
}

@inproceedings{song20192d,
  title={2d lidar map prediction via estimating motion flow with gru},
  author={Song, Yafei and Tian, Yonghong and Wang, Gang and Li, Mingyang},
  booktitle={2019 International Conference on Robotics and Automation (ICRA)},
  pages={6617--6623},
  year={2019},
  organization={IEEE}
}

@article{mahjourian2022occupancy,
  title={Occupancy flow fields for motion forecasting in autonomous driving},
  author={Mahjourian, Reza and Kim, Jinkyu and Chai, Yuning and Tan, Mingxing and Sapp, Ben and Anguelov, Dragomir},
  journal={IEEE Robotics and Automation Letters},
  volume={7},
  number={2},
  pages={5639--5646},
  year={2022},
  publisher={IEEE}
}

@inproceedings{liu2025let,
  title={Let Occ Flow: Self-Supervised 3D Occupancy Flow Prediction},
  author={Liu, Yili and Mou, Linzhan and Yu, Xuan and Han, Chenrui and Mao, Sitong and Xiong, Rong and Wang, Yue},
  booktitle={Conference on Robot Learning},
  pages={2895--2912},
  year={2025},
  organization={PMLR}
}

@inproceedings{leng2025occupancy,
  title={Occupancy learning with spatiotemporal memory},
  author={Leng, Ziyang and Yang, Jiawei and Yi, Wenlong and Zhou, Bolei},
  booktitle={Proceedings of the IEEE/CVF International Conference on Computer Vision},
  pages={26569--26578},
  year={2025}
}

@article{mohan2026forecastocc,
  title={ForecastOcc: Vision-based Semantic Occupancy Forecasting},
  author={Mohan, Riya and Hurtado, Juana Valeria and Mohan, Rohit and Valada, Abhinav},
  journal={arXiv preprint arXiv:2602.08006},
  year={2026}
}

@INPROCEEDINGS{LangeB-RSS-25, 
    AUTHOR    = {Bernard Lange AND Masha Itkina AND Jiachen Li AND Mykel Kochenderfer}, 
    TITLE     = {{Self-supervised Multi-future Occupancy Forecasting for Autonomous Driving}}, 
    BOOKTITLE = {Proceedings of Robotics: Science and Systems}, 
    YEAR      = {2025}, 
    DOI       = {10.15607/RSS.2025.XXI.003} 
}

@inproceedings{wang2025diffusion,
  title={Diffusion-Based Generative Models for 3D Occupancy Prediction in Autonomous Driving},
  author={Wang, Yunshen and Liu, Yicheng and Yuan, Tianyuan and Mao, Yucheng and Liang, Yingshi and Yang, Xiuyu and Zhang, Honggang and Zhao, Hang},
  booktitle={2025 IEEE International Conference on Robotics and Automation (ICRA)},
  pages={8322--8328},
  year={2025},
  organization={IEEE}
}

@inproceedings{xie2023stochastic,
  title={Stochastic occupancy grid map prediction in dynamic scenes},
  author={Xie, Zhanteng and Dames, Philip},
  booktitle={Conference on Robot Learning},
  pages={1686--1705},
  year={2023},
  organization={PMLR}
}

@article{xie2025scope,
  title={Scope: Stochastic cartographic occupancy prediction engine for uncertainty-aware dynamic navigation},
  author={Xie, Zhanteng and Dames, Philip},
  journal={IEEE Transactions on Robotics},
  year={2025},
  publisher={IEEE}
}

@inproceedings{trautman2010unfreezing,
  title={Unfreezing the robot: Navigation in dense, interacting crowds},
  author={Trautman, Peter and Krause, Andreas},
  booktitle={2010 IEEE/RSJ International Conference on Intelligent Robots and Systems},
  pages={797--803},
  year={2010},
  organization={IEEE}
}

@inproceedings{cao2019dynamic,
  title={Dynamic channel: A planning framework for crowd navigation},
  author={Cao, Chao and Trautman, Peter and Iba, Soshi},
  booktitle={2019 international conference on robotics and automation (ICRA)},
  pages={5551--5557},
  year={2019},
  organization={IEEE}
}

@article{mavrogiannis2023winding,
  title={Winding Through: Crowd Navigation via Topological Invariance},
  author={Mavrogiannis, Christoforos and Balasubramanian, Krishna and Poddar, Sriyash and Gandra, Anush and Srinivasa, Siddhartha S},
  journal={IEEE Robotics and Automation Letters},
  volume={8},
  number={1},
  pages={121--128},
  year={2023}
}

@inproceedings{zhang2025ga,
  title={GA-TEB: Goal-Adaptive Framework for Efficient Navigation Based on Goal Lines},
  author={Zhang, Qianyi and Luo, Wentao and Zhang, Ziyang and Wang, Yaoyuan and Liu, Jingtai},
  booktitle={2025 IEEE International Conference on Robotics and Automation (ICRA)},
  pages={3876--3882},
  year={2025},
  organization={IEEE}
}

@inproceedings{de2023globally,
  title={Globally Guided Trajectory Planning in Dynamic Environments},
  author={de Groot, OM and Ferranti, L and Gavrila, D and Alonso-Mora, J},
  booktitle={ICRA 2023: International Conference on Robotics and Automation},
  pages={10118--10124},
  year={2023},
  organization={IEEE}
}

@article{de2025topology,
  title={Topology-Driven Parallel Trajectory Optimization in Dynamic Environments},
  author={de Groot, Oscar and Ferranti, Laura and Gavrila, Dariu M and Alonso-Mora, Javier},
  journal={IEEE Transactions on Robotics},
  volume={41},
  pages={110--126},
  year={2025}
}

@article{thrun2003learning,
  title={Learning occupancy grid maps with forward sensor models},
  author={Thrun, Sebastian},
  journal={Autonomous robots},
  volume={15},
  number={2},
  pages={111--127},
  year={2003},
  publisher={Springer}
}

@article{an2023latent,
  title={Latent-shift: Latent diffusion with temporal shift for efficient text-to-video generation},
  author={An, Jie and Zhang, Songyang and Yang, Harry and Gupta, Sonal and Huang, Jia-Bin and Luo, Jiebo and Yin, Xi},
  journal={arXiv preprint arXiv:2304.08477},
  year={2023}
}

@inproceedings{salimans2022progressive,
title={Progressive Distillation for Fast Sampling of Diffusion Models},
author={Tim Salimans and Jonathan Ho},
booktitle={International Conference on Learning Representations},
year={2022},
url={https://openreview.net/forum?id=TIdIXIpzhoI}
}

@inproceedings{lu2024vdt,
  title={Vdt: General-purpose video diffusion transformers via mask modeling},
  author={Lu, Haoyu and Yang, Guoxing and Fei, Nanyi and Huo, Yuqi and Lu, Zhiwu and Luo, Ping and Ding, Mingyu},
  booktitle={International Conference on Learning Representations},
  volume={2024},
  pages={19259--19286},
  year={2024}
}

@article{fox2002dynamic,
  title={The dynamic window approach to collision avoidance},
  author={Fox, Dieter and Burgard, Wolfram and Thrun, Sebastian},
  journal={IEEE robotics \& automation magazine},
  volume={4},
  number={1},
  pages={23--33},
  year={2002},
  publisher={IEEE}
}

@article{lee2025adaptive,
  title={Adaptive Trajectory Refinement for Optimization-based Local Planning in Narrow Passages},
  author={Lee, Hahjin and Kim, Young J},
  journal={arXiv preprint arXiv:2510.26142},
  year={2025}
}

\end{document}